\documentclass[conference]{IEEEtran}
\usepackage{cite}
\ifCLASSINFOpdf
  \usepackage[pdftex]{graphicx}
  \graphicspath{{./figures/}}
\else
  \usepackage[dvips]{graphicx}
\fi
\usepackage{amsmath}
\usepackage{algorithmic}
\usepackage{array}
\usepackage{url}
\usepackage{xcolor}
\usepackage{booktabs,amsfonts}
\usepackage[nolist,nohyperlinks]{acronym}

\makeatletter
\def\ps@IEEEtitlepagestyle{
  \def\@oddfoot{\mycopyrightnotice}
  \def\@evenfoot{}
}
\def\mycopyrightnotice{
  {\footnotesize
  \begin{minipage}{\textwidth}
  \copyright~2020 IEEE.  Personal use of this material is permitted.  Permission from IEEE must be obtained for all other uses, in any current or future media, including reprinting/republishing this material for advertising or promotional purposes, creating new collective works, for resale or redistribution to servers or lists, or reuse of any copyrighted component of this work in other works.
  \end{minipage}
  }
}
\begin{document}
\title{Memory Organization for Energy-Efficient Learning and Inference in Digital Neuromorphic Accelerators}

\author{\IEEEauthorblockN{ Clemens JS Schaefer\textsuperscript{*}, Patrick Faley\textsuperscript{*}, Emre O Neftci\textsuperscript{$\dagger$} and Siddharth Joshi\textsuperscript{*}}
\IEEEauthorblockA{\textsuperscript{*}Department of Computer Science and Engineering, University of Notre Dame, Notre Dame, IN, USA } 

\IEEEauthorblockA{\textsuperscript{$\dagger$}Department of Cognitive Sciences and Department of Computer Science, UC Irvine, Irvine, CA, USA}
Email: \{cschaef6, pfaley, sjoshi2\}@nd.edu, eneftci@uci.edu
}

\maketitle

\IEEEpeerreviewmaketitle

\begin{abstract}
The energy efficiency of neuromorphic hardware is greatly affected by the energy of storing, accessing, and updating synaptic parameters. Various methods of memory organisation targeting energy-efficient digital accelerators have been investigated in the past, however, they do not completely encapsulate the energy costs at a system level. To address this shortcoming and to account for various overheads, we synthesize the controller and memory for different encoding schemes and extract the energy costs from these synthesized blocks. Additionally, we  introduce  functional encoding for structured connectivity such  as  the  connectivity  in  convolutional layers. Functional encoding offers a 58\% reduction in the energy to implement a backward pass and weight update in such layers compared to existing index-based solutions. We show that for a 2 layer spiking neural network trained to retain a spatio-temporal pattern, bitmap (PB-BMP) based organization can encode the sparser networks more efficiently. This form of encoding delivers a 1.37$\times$ improvement in energy efficiency coming at the cost of a 4\% degradation in network retention accuracy as measured by the van Rossum distance.
\end{abstract}

\section{Introduction}
\acrodef{AC}[AC]{Arrenhius \& Current}
\acrodef{ANN}[ANN]{Artificial Neural Network}
\acrodef{AER}[AER]{Address Event Representation}
\acrodef{AEX}[AEX]{AER EXtension board}
\acrodef{AMDA}[AMDA]{``AER Motherboard with D/A converters''}
\acrodef{API}[API]{Application Programming Interface}
\acrodef{BP}[BP]{Back-Propagation}
\acrodef{BPTT}[BPTT]{Back-Propagation-Through-Time}
\acrodef{BM}[BM]{Boltzmann Machine}
\acrodef{CAVIAR}[CAVIAR]{Convolution AER Vision Architecture for Real-Time}
\acrodef{CCN}[CCN]{Cooperative and Competitive Network}
\acrodef{CD}[CD]{Contrastive Divergence}
\acrodef{CMOS}[CMOS]{Complementary Metal--Oxide--Semiconductor}
\acrodef{COTS}[COTS]{Commercial Off-The-Shelf}
\acrodef{CPU}[CPU]{Central Processing Unit}
\acrodef{CV}[CV]{Coefficient of Variation}
\acrodef{CV}[CV]{Coefficient of Variation}
\acrodef{DAC}[DAC]{Digital--to--Analog}
\acrodef{DBN}[DBN]{Deep Belief Network}
\acrodef{DCLL}[DECOLLE]{Deep Continuous Local Learning}
\acrodef{DFA}[DFA]{Deterministic Finite Automaton}
\acrodef{DFA}[DFA]{Deterministic Finite Automaton}
\acrodef{divmod3}[DIVMOD3]{divisibility of a number by 3}
\acrodef{DPE}[DPE]{Dynamic Parameter Estimation}
\acrodef{DPI}[DPI]{Differential-Pair Integrator}
\acrodef{DSP}[DSP]{Digital Signal Processor}
\acrodef{DVS}[DVS]{Dynamic Vision Sensor}
\acrodef{EDVAC}[EDVAC]{Electronic Discrete Variable Automatic Computer}
\acrodef{EIF}[EI\&F]{Exponential Integrate \& Fire}
\acrodef{EIN}[EIN]{Excitatory--Inhibitory Network}
\acrodef{EPSC}[EPSC]{Excitatory Post-Synaptic Current}
\acrodef{EPSP}[EPSP]{Excitatory Post--Synaptic Potential}
\acrodef{eRBP}[eRBP]{Event-Driven Random Back-Propagation}
\acrodef{FPGA}[FPGA]{Field Programmable Gate Array}
\acrodef{FSM}[FSM]{Finite State Machine}
\acrodef{GPU}[GPU]{Graphical Processing Unit}
\acrodef{HAL}[HAL]{Hardware Abstraction Layer}
\acrodef{HH}[H\&H]{Hodgkin \& Huxley}
\acrodef{HMM}[HMM]{Hidden Markov Model}
\acrodef{HW}[HW]{Hardware}
\acrodef{hWTA}[hWTA]{Hard Winner--Take--All}
\acrodef{IF2DWTA}[IF2DWTA]{Integrate \& Fire 2--Dimensional WTA}
\acrodef{IF}[I\&F]{Integrate \& Fire}
\acrodef{IFSLWTA}[IFSLWTA]{Integrate \& Fire Stop Learning WTA}
\acrodef{INCF}[INCF]{International Neuroinformatics Coordinating Facility}
\acrodef{INI}[INI]{Institute of Neuroinformatics}
\acrodef{IO}[IO]{Input-Output}
\acrodef{IPSC}[IPSC]{Inhibitory Post-Synaptic Current}
\acrodef{ISI}[ISI]{Inter--Spike Interval}
\acrodef{JFLAP}[JFLAP]{Java - Formal Languages and Automata Package}
\acrodef{LIF}[LI\&F]{Linear Integrate \& Fire}
\acrodef{LSM}[LSM]{Liquid State Machine}
\acrodef{LTD}[LTD]{Long-Term Depression}
\acrodef{LTI}[LTI]{Linear Time-Invariant}
\acrodef{LTP}[LTP]{Long-Term Potentiation}
\acrodef{LTU}[LTU]{Linear Threshold Unit}
\acrodef{MCMC}{Markov Chain Monte Carlo}
\acrodef{NHML}[NHML]{Neuromorphic Hardware Mark-up Language}
\acrodef{NMDA}[NMDA]{NMDA}
\acrodef{NME}[NE]{Neuromorphic Engineering}
\acrodef{PCB}[PCB]{Printed Circuit Board}
\acrodef{PRC}[PRC]{Phase Response Curve}
\acrodef{PSC}[PSC]{Post-Synaptic Current}
\acrodef{PSP}[PSP]{Post--Synaptic Potential}
\acrodef{RI}[KL]{Kullback-Leibler}
\acrodef{RRAM}[RRAM]{Resistive Random-Access Memory}
\acrodef{RBM}[RBM]{Restricted Boltzmann Machine}
\acrodef{ROC}[ROC]{Receiver Operator Characteristic}
\acrodef{SAC}[SAC]{Selective Attention Chip}
\acrodef{SCD}[SCD]{Spike-Based Contrastive Divergence}
\acrodef{SCX}[SCX]{Silicon CorteX}
\acrodef{SRM}[SRM]{Spike Response Model}
\acrodef{SNN}[SNN]{Spiking Neural Network}
\acrodef{STDP}[STDP]{Spike Time Dependent Plasticity}
\acrodef{SW}[SW]{Software}
\acrodef{sWTA}[SWTA]{Soft Winner--Take--All}
\acrodef{VHDL}[VHDL]{VHSIC Hardware Description Language}
\acrodef{VLSI}[VLSI]{Very  Large  Scale  Integration}
\acrodef{WTA}[WTA]{Winner--Take--All}
\acrodef{XML}[XML]{eXtensible Mark-up Language}

Biological organisms operate autonomously with extreme energy efficiency, learning continuously amid unreliable and noisy environmental stimuli. Designing artificial autonomous systems that can embody these traits remains a grand challenge in engineering. Taking inspiration from biology, autonomous systems using biologically inspired computational and sensory systems have been able to deliver remarkable results over the past few decades~\cite{delbruck2013robotic,sironi2018hats,vidal2018ultimate,kaiser2019embodied}. Neuromorphic computing aims to develop hardware and algorithms that embody the principles upon which biology operates~\cite{Qiao_etal15_recoon-l, Merolla_etal14_millspik, Schemmel_etal10_wafeneur}, crucially, while maximizing the energy efficiency of learning.

Neuromorphic hardware platforms designed to efficiently run large-scale spiking neural networks generally consist of a neuro-synaptic core and a communication fabric to connect multiple such cores. The neurosynaptic core, in turn, is generally composed of some form of a neuron subsystem which implements spike accumulation and all calculations associated with membrane potential dynamics; a synapse subsystem which stores associated weights and implements synaptic dynamics; and if learning is supported, then additional circuitry to implement synaptic plasticity~\cite{Merolla_etal14_millspik,Davies_etal18_loihneur,pei2019Tianjic,Qiao_etal15_recoon-l,Friedmann_etal17_demohybr}. 

In this paper, we focus on representations of synaptic connectivity in digital memories for energy-efficient inference and learning in neuromorphic spiking architectures.
%\sj{is this still valid? Specifically, we investigate three-factor rules, which can be viewed as extensions of Hebbian learning and \ac{STDP} with an external modulatory signal.} Such rules are capable of a wide number of unsupervised, supervised, and reinforcement learning paradigms, and implementations can have scaling properties comparable to that of \ac{STDP} \cite{Payvand_etal19_errothre}. Three-factor rules can be derived from first principles \cite{Urbanczik_Senn14_learby} such as gradient descent on a global loss function. 
Specifically, we examine the energetic impact of implementing  backpropagation-through-time (BPTT) using a spike-based learning rule together with a surrogate gradient (SG). This learning rule minimizes a global loss function, together with the SG, BPTT can enable supervised gradient based learning despite the discontinuity induced by spiking non-linearities encountered in \acp{SNN}~\cite{Neftci_etal19_surrgrad}.  Over the course of this paper, we examine how different storage schemes impact the energy required to implement BPTT.

In digital memories implemented in large-scale neuromorphic systems, the organization and representation of synaptic parameters in memory impacts the energy efficiency of learning~\cite{kim2018efficient} and inference~\cite{aimar2018nullhop}. This is exemplified in the difference in forward and reverse memory access patterns in a crossbar memory and an index-based memory~\cite{Pedroni_etal16_forwtabl}. In a crossbar memory, since both the synaptic weight matrix and its transpose are immediately accessible~\cite{Payvand_etal19_errothre}, given a pre-synaptic neuron all post-synaptic neurons connected to it can be determined and with equal ease, given a post-synaptic neuron, all neurons pre-synaptic to it can be determined. However, index-based storage schemes often sacrifice ease of reverse access for storage efficiency. This has led to investigations into various formats for storage efficiency~\cite{joshi2017neuromorphic, pedroni2019memory, kim2018efficient}. Different structures for forward access storage efficiency are examined in ~\cite{joshi2017neuromorphic}, which argue for a bitmap based approach. This was countered by investigations in ~\cite{pedroni2019memory} which determined that storage and complexity of access are both crucial, arguing for an indexed-list based approach. However, rather than directly studying energy efficiency, prior work has examined this question through the lens of number of bits needed for storage or the number of operations required to access the weights. This paper explores the design space of memory organization against the more direct metric of access energy in order to re-examine the results of prior work in light of both algorithmic and circuit considerations. 

The central contributions of this paper are two-fold: first, we introduce a functional approach to storing regular and structured connectivity such as the connectivity in convolutional layers; second, rather than taking access efficiency or storage as proxies for energy as done in~\cite{joshi2017neuromorphic, pedroni2019memory}, we directly compare the energy cost induced by the different data encoding schemes. In order to accomplish this energy comparison, we obtain the energy of the different schemes by synthesizing a datapath, a controller, and the associated memories in a 40~nm CMOS technology. The memory results were verified against CACTI~\cite{balasubramonian2017cacti} (a SPICE accurate architectural memory model) to ensure correctness. 

\section{Background}

\subsection{Surrogate Gradient Learning}\label{sec:SG}
The model used in this paper consists of networks of plastic integrate-and-fire neurons, expressed here in discrete time: 
\begin{equation}\label{eq:lif_equations}
  \begin{split}
    U_i^{(l)}[n] &= \sum_j W_{ij}^{(l)} P_j[n] - \delta R_i[n], \\
  \end{split}
\end{equation}
\[
  \begin{split} 
    S_i^{(l)}[n] &= \Theta( U_i^{(l)}[n]-\vartheta)\\
    Q_j[n+1] &= \alpha  Q_{j}[n] + S_j^{(l-1)}[n], \\  
    P_j[n+1] &= \beta P_{j}[n] + Q_{j}[n], \\
    R_i[n+1] &= \gamma R_{i}[n] + S_{i}[n]. \\
  \end{split}
\]
where $U_i^{(l)}[n]$ is the membrane potential of neuron $i$ at layer $l$ at time step $n$, $W$ is synaptic weight matrix, $\vartheta$ is the firing threshold and $S_i^{(l)}$ is the spiking output of this neuron. 
The function $\Theta$ is the step function, \emph{i.e.} $S_i^{(l)}[n]=1$ when ${U_i^{(l)}[n]=0}$. The constants $\beta$, $\gamma$ and $\alpha$ capture the decay dynamics of the membrane potential $U_i$, the refractory (resetting) state $R_i$ and the synaptic state $Q_i$ and can be related to time constants in leaky integrate-and-fire neurons.
States $P$ and $Q$ describe the traces of the membrane and the current-based synapse, respectively. 
$R$ is a refractory state that resets and inhibits the neuron after it has emitted a spike, and $\delta$ is the constant that controls its magnitude.
Note that \eqref{eq:lif_equations} is equivalent to a discrete-time version of the \ac{SRM}$_0$ with linear filters \cite{Gerstner_Kistler02_spikneur}.
This \ac{SNN} and the ensuing learning dynamics can be transformed into a standard binary neural network by setting all $\alpha=0$, replacing all $P_j[n]$ with ${S_j^{(l)}[n-1]}$ and dropping $Q$ and $R$.

Assuming a global cost function $\mathcal{L}$, the gradients with respect to the weights in layer $l$ can be computed using backpropagation-through-time \cite{Neftci_etal19_surrgrad}.
$\Theta$ is non-differentiable but following a surrogate gradient learning, $\Theta$'s derivative can be replaced by a smooth sigmoidal or piecewise constant function for optimization purposes \cite{Neftci_etal19_surrgrad}.
Our experiments make use of the normalized negative part of a fast sigmoid function and a Van Rossum distance \cite{Zenke_Ganguli17_supesupe}.

\subsection{Quantization}
\begin{figure}[!t]
\centering
\includegraphics[width=.88\columnwidth]{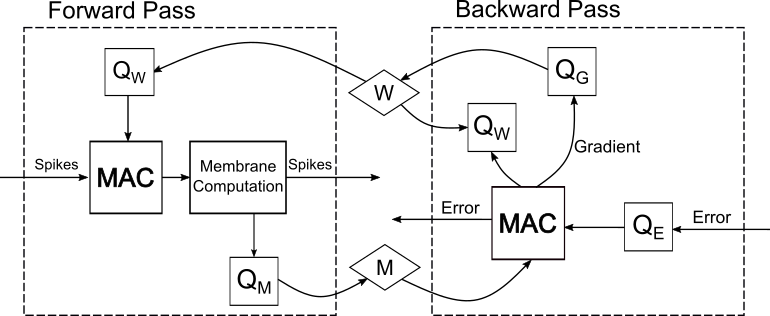}
\caption{Quantization schematic illustrating the quantization process in forward and backward pass. Note the membrane potential (M) needs to be stored for the backward pass. Squares indicate operations (i.e. $Q_E$ quantize error) and diamonds stored values (i.e. membrane potential and weights).} 
\label{fig:sim}
\end{figure}

Efficient implementations of learning on-chip entail learning with quantized weights, gradients, and membrane voltage dynamics. To accurately model the effect of quantizing the weight and gradient values we follow the procedures outlined in~\cite{wu2018training} and~\cite{yousefzadeh2018practical}, as shown in Fig.~\ref{fig:sim}. Weights are quantized by restricting them to a feasible weight range defined by ($\text{min}~=~-1~+~\sigma(b_w)$ and  $\text{max}~=~+1~-~\sigma(b_w)$), where $b_w$ is the number of bits encoding the weight and $\sigma(b) =  2^{1-b}$.

To prevent overflow, weights in each layer are scaled by $\eta$ where:
\begin{equation}
    \eta = 2^{ \text{round} \left[ \log_2 \left(\frac{(\frac{1}{\sigma(b_w)}-0.5)\cdot \sigma(b_w) }{\sqrt{\frac{3}{\text{fan in}}}}  \right) \right]},
\end{equation}
and \textit{fan in} represents the number of connections into a layer. % neurons presynaptic to a given neuron.
In the backward pass the error is first normalized by its greatest absolute value and then clipped and quantized to ensure precision is maintained. After computing and normalizing the gradients, stochastic rounding is applied to increase the gradient precision when learning over multiple epochs.
\subsection{Encoding Connectivity and Weights}
%Our Weight Accessing Method, bits required etc
\begin{figure}[!t]
\centering
\includegraphics[width=.88\columnwidth]{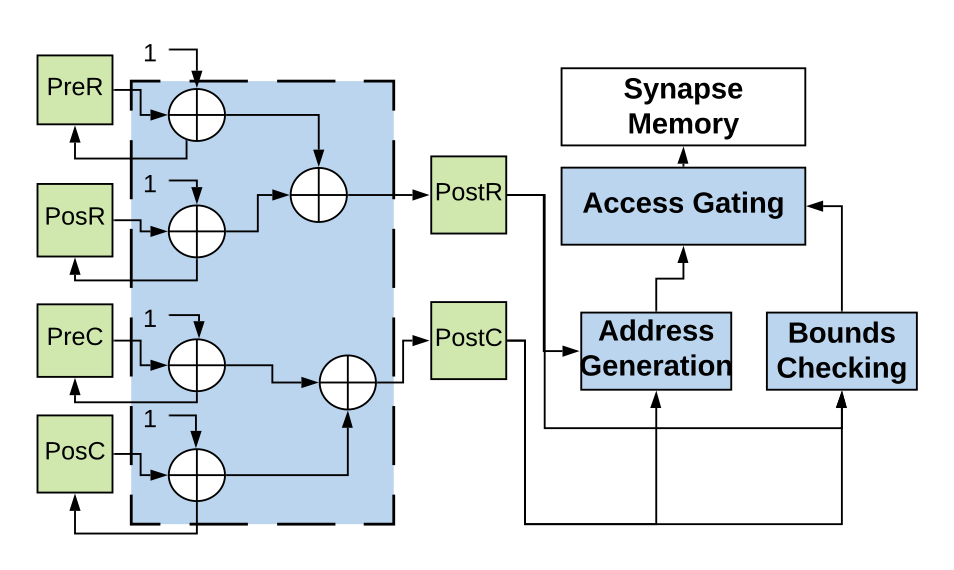}
\caption{The schematic of the logic used to generate the convolutional sparsity pattern. The blue-shaded elements are combinational logic while the green shaded elements are registers. Precomputed parameters are used to determine if the read address is valid, preventing spurious reads to minimize energy consumption.}
\label{fig:forwardfunc}
\end{figure}

Index based and bitmap based representations can improve the efficiency with which different synaptic connectivity patterns are stored in digital memories~\cite{jin2010stdp:spinnaker,aimar2018nullhop,joshi2017neuromorphic, pedroni2019memory}. Here, we very briefly introduce the terminology used in this paper, but refer the readers to~\cite{pedroni2019memory,joshi2017neuromorphic} for a more in-depth review of the topic. Crossbar (CB) based storage schemes sequentially store all potential synaptic parameters between input (pre-synaptic neurons) and outputs (post-synaptic neurons) in the memory. This offers constant-time access to any synapse based on the post and pre-synaptic neuron address~\cite{Merolla_etal11_digineur,kim2018efficient}.

Alternatively, index based methods are better suited to storing sparse connections. Such methods only store the nonzero connections and an additional set of pointers, obviating the need to store any absent synapses. %There a different implementation of this method, the index based bitmap storage method keeps track of the connectivity structure with a bitmap and a row pointer list for the actual weight addresses. Thus requiring a memory budget of $MN + \lceil \rho N \rceil Mk + M \log_2 (\lceil \rho N\rceil)$~\cite{joshi2017neuromorphic}.
Two sparse representation schemes have been introduced in~\cite{joshi2017neuromorphic,pedroni2019memory} within the context of weight storage; these are the compressed sparse row (PB-CSR) and pointer based bitmap (PB-BMP). Due to the flexibility afforded by different index-based sparse storage schemes they are also employed in Intel Loihi~\cite{Davies_etal18_loihneur}.
In addition to these general representation schemes, this paper also introduces a functional scheme to better represent regular and patterned connectivity. Through functional encoding, the connectivity information can be derived through run-time reconfigurable  combinational logic shown in Fig.~\ref{fig:forwardfunc}. Hence only the synaptic parameters that exist need be stored. Zero-weight synapses don't require explicit storage and the connectivity can be computed through the function. Since, the connectivity is computed rather than stored, this saves on the memory size as well as the energy required per memory access. This saves on both the size of the memory and the number of reads to the memory which are more expensive than evaluation combinational logic.

Encoding the connectivity pattern induced by convolutions into a function (see~\eqref{eq:forward},~\eqref{eq:back}) incurs a few integer additions and subtractions while iterating over the size of the convolutional kernel. The latency and energy cost of this is minimal in comparison to a register file look-up. Further, parts of this can be executed in parallel to enable multi-bank memory access if required. The function that governs the forward lookup is written as a two-dimensional function given by:
\begin{equation}\label{eq:forward}
   \begin{split}
    f(PreR, PosR) &= PreR + PosR\\
    f(PreC,PosC) &= PreC + PosC
    \end{split}
\end{equation}    
where, \textit{Pre} and \textit{Post}, represent the pre and post-synaptic neurons respectively, \textit{R} and \textit{C} represent the row and column in a 2D kernel, and \textit{pos} represents an iterator to generate all fanouts for a given presynaptic neuron. Similarly, the inverse function is given by:

\begin{equation}\label{eq:back}
  \begin{split}
    f^{-1}(PostR,PosR) &= PostR - PosR\\ 
    f^{-1}(PostC,PosC) &= PostC - PosC.
  \end{split}
\end{equation}

\section{Energy Impact of Connectivity Encoding}
% \begin{figure}[!t]
% \centering
% \fbox{\includegraphics[width=.88\columnwidth]{figures/fc.png}}
% \caption{Energy trade-offs for different levels of weight quantization for a fully connected layer (input 728, output 128) given a sparsity level of 25\%. The top panel shows the energy cost associated with memory accesses during a forward pass and bottom panel shows the energy cost associated with memory access and writes during a backward pass.} 
% \label{fig_fc}
% \end{figure}

% \begin{figure}[!t]
% \centering
% \includegraphics[width=.88\columnwidth]{figures/conv.png}
% \caption{Energy trade-offs for different levels of weight quantization for a convolutional layer (input $28\times28$, size $3\times3$, $32$ in channels, $32$ out channels). Top panel shows energy cost incurred during a forward pass and bottom panel shows the energy cost incurred a backward pass.}
% \label{fig_conv}
% \end{figure}

\begin{figure}[!t]

\centering
\includegraphics[width=1\columnwidth]{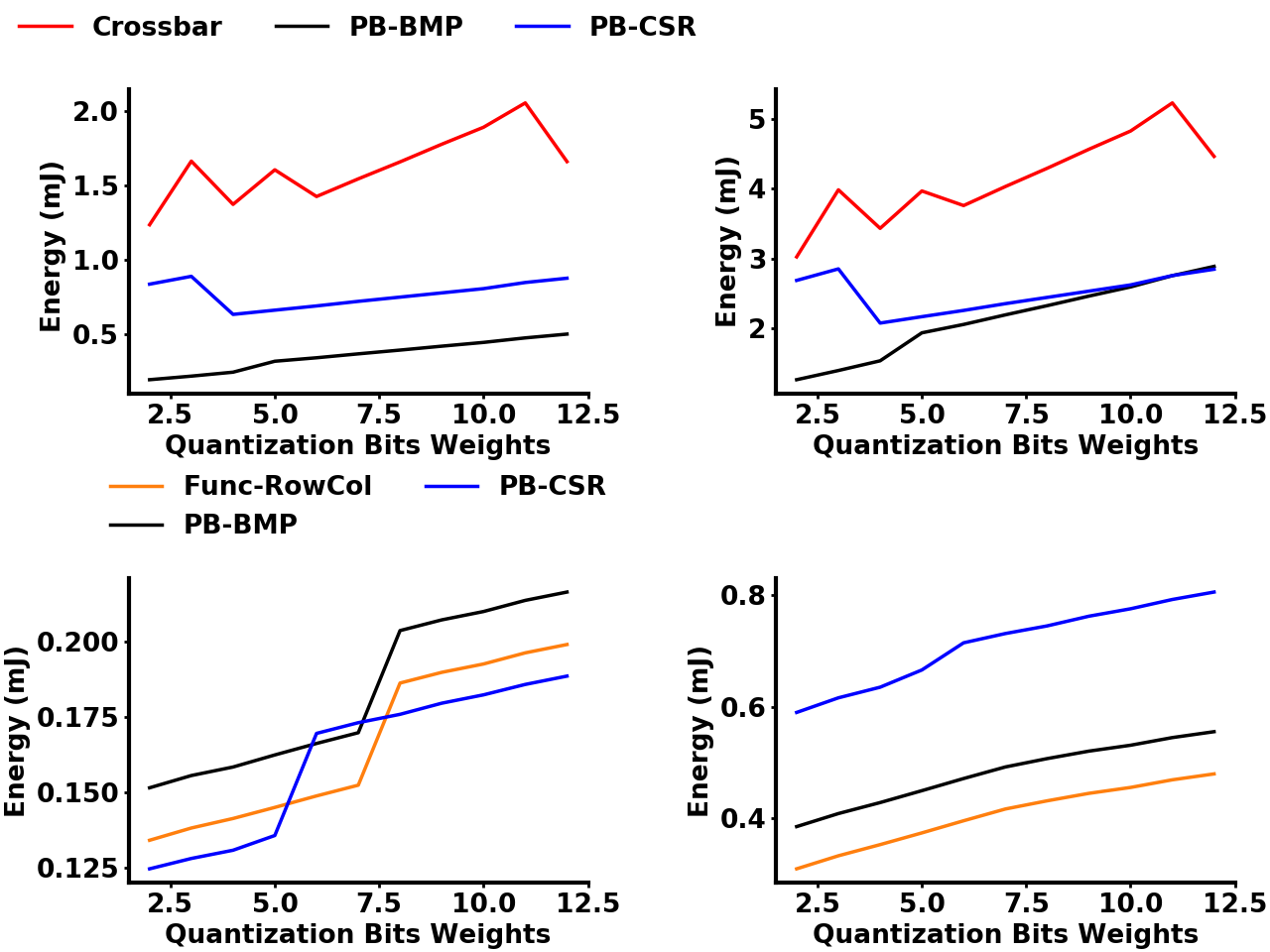}
\caption{Top: Energy trade-offs for different levels of weight quantization for a fully connected layer (input 728, output 128) given a sparsity level of 25\%. The left graph shows the energy cost associated with memory accesses during a forward pass and right graph shows the energy cost associated with memory access and writes during a backward pass. Bottom: Energy trade-offs for different levels of weight quantization for a convolutional layer (input $28\times28$, size $3\times3$, $32$ in channels, $32$ out channels). Left graph shows energy cost incurred during a forward pass and right graph shows the energy cost incurred a backward pass.}  % \cs{I dont think the 4 panel idea buys us anything. Old version still in latex code.}
\label{fig_fc}
%\label{fig_conv}
\end{figure}

\begin{figure}[!t]
\centering
\includegraphics[width=.88\columnwidth]{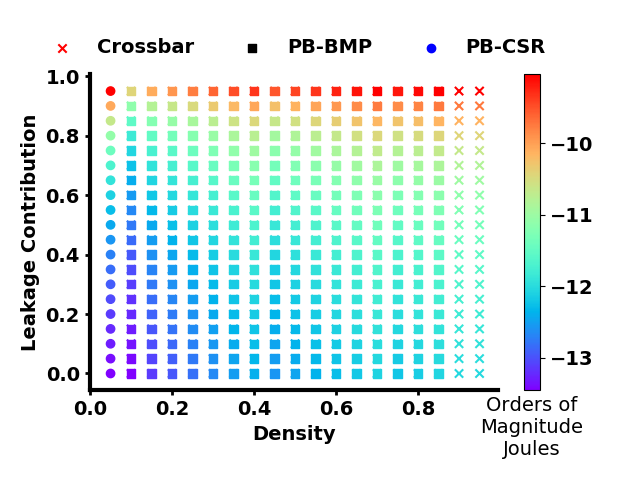}
\caption{Order of magnitude of the minimal energy consumption for a fully connected layer (input 728, output 128) with 8-bit weight precision. For the range of leakage energy contribution and the density nonzeroes in the layer, we denote the energy through the color and the most efficient storage scheme through the symbol. The energy is calculated for a combination of the forward and backward pass.}
\label{fig_sparsity}
\end{figure}

In order to study the interplay between memory storage, access efficiency, synaptic parameter quantization, and energy efficiency we implement these storage schemes at the RTL level.  We synthesized controllers and the memory and datapaths, for the different encoding strategies, in 40~nm CMOS using a standard Synopsys flow. Memory read and write energies extracted from this are verified against a CACTI model for the same technology-node. In all the experiments reported below, unless otherwise mentioned, only the active energy of this synthesized system is reported. These energies are reported for one entire forward or backward pass of a layer in an \ac{SNN}. In each case the indirection tables and the weight tables are split into different memories, to minimize the energy-cost of indirection. Figure~\ref{fig_fc} shows the effect of weight precision on the energy efficiency for different encoding schemes for both the convolution (bottom) and the fully connected (top) layers. The energy for the forward pass across an entire layer is shown on the left, while the energy for the backward pass for an entire layer is shown on the right. We examine the energy required to implement a forward and a backward pass for a fully connected layer of size $728\times 128$ with a density of 75\% (sparsity of 25\%). For the FC layer, in the forward pass, the PB-BMP structure delivers the lowest energy over a range of weight quantization values. This is due to the more compact representation leading to smaller memories which are not penalized as much during write operations. For the convolutional layer, we implement convolution over an input of size $28\times28$ with a filter of size $3\times3$. While the energy cost of calculating the connectivity through a function is comparable to that of the index based PB-CSR, the energy consumed in the backward pass when using our function is much lower due to far fewer memory accesses.

Figure~\ref{fig_sparsity} shows the performance of the different connectivity encoding schemes when connection density is varied over the range 5\% to 100\% while simultaneously varying the sparsity of input activity. Since the frequency of activity corresponds to the frequency with which a forward and reverse pass through the memory occur, sparse activity would correspond to leakage power from the memory dominating. Interestingly, the results contradict general assumption from previous work~\cite{pedroni2019memory} where the number of memory accesses dominated all energy. This is because the size of the memory has an out-sized impact on the access and write energy as well as the leakage energy.

\section{System Design Considerations}

\begin{figure}[!t]
\centering
\includegraphics[width=.88\columnwidth]{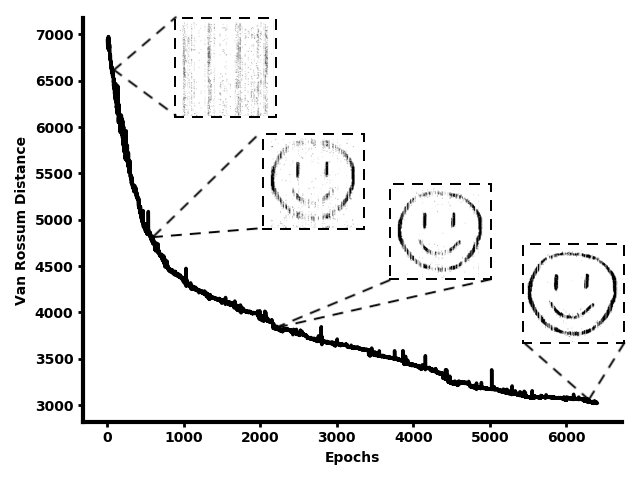}
\caption{A learning curve with an exemplary output of the SNNs after 60, 600, 2000 and 6400 epochs, we use the Van Rossum distance as a measure of the SNNs performance.}
\label{fig_vr_meaning}
\end{figure}

We analyze the effect of different encoding schemes and varying resolution on the accuracy of a spiking network trained as outlined in Section~\ref{sec:SG}. The network, trained on a spatio-temporal pattern as shown in Fig.~\ref{fig_vr_meaning},  consists of one input layer with 700 neurons, one hidden layer with 400 neurons, and 250 output neurons. We provide an input of 700 Poisson spike trains over 250 time steps with varying inter-spike intervals. The target was generated by taking a clean pattern and multiplying it with Bernoulli noise ($p=.95$). We use the van Rossum (VR) distance~\cite{rossum2001novel} as a loss function; the VR distance calculations are performed at floating point precision and the energy for these calculations is ignored in the experiments. The parameters defining neuron and synaptic dynamics are set values amenable to compact hardware, \textit{e.g.} single-tap FIR filters for membrane and synaptic dynamics. The VR distance is then recorded over $10000$ epochs of training.

\begin{figure}[!t]
\centering
\includegraphics[width=.88\columnwidth]{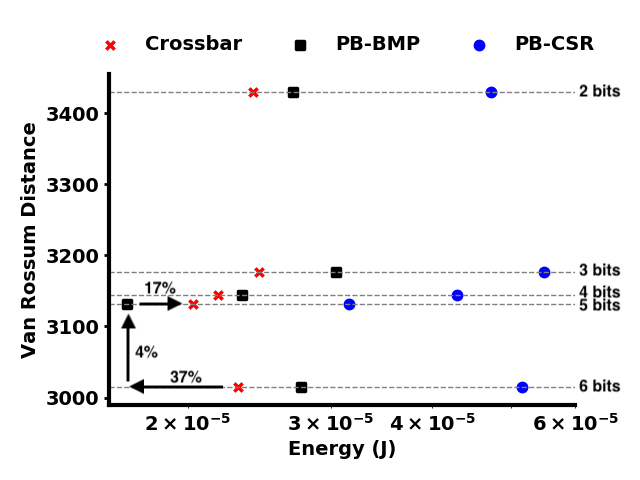}
\caption{The accuracy and energy tradeoff for a 700-400-250 neural network at different weight precisions. We annotate the accuracy achieved by a network with the respective quantization levels: 2-6 bits. The weights in the network are encoded using CB, PB-BMP, and PB-CSR memory access schemes, since lower precision also results in a larger number of zeros and thus greater sparsity, this impacts the efficiency with which the that network can be processed.}
\label{fig_frontier}
\end{figure}

We determine the energy required to encode the two layers of the \ac{SNN}  through the different encoding schemes. Different bit-precisions affect the accuracy (VR distance) achieved by the network while simultaneously affecting the energy associated with accessing and writing to the weight memories. We capture these trade-offs in Fig.~\ref{fig_frontier}. The highest accuracy is achieved with 6 bit weight resolution, where the lowest energy consumption corresponds to the CB structure. However, the PB-BMP scheme, when used with 5-bit quantized weights is the most energetically-efficient scheme across all experiments. This is in part due to the network having a sparsity of 27\% over the course of the training.

% \vspace{-.15cm}

\section{Conclusion}
To summarize, we proposed a new functional method to encode connectivity and weight storage for convolutional layers. We tested both fully-connected and convolutional layers and analyzed the impact of the storage cost as well as weight access cost on the net energy. This provides a more holistic view of the design of digital systems implementing \acp{SNN}, unlike previous work~\cite{joshi2017neuromorphic, pedroni2019memory} which focused mostly on storage cost. Using the proposed function to encode structured connectivity approximately doubles the energy efficiency of implementing the backward pass of a convolutional layer composed of 8-bit synapses when compared to the PB-CSR. In the forward pass, this function incurs an overhead of $5\%$ in energy over the PB-CSR. Leading to a net energy saving when weight updates are frequent, such as in the context of continuous learning. Additionally, we compared the energy required to train a quantized SNN stored using various data encoding schemes. An SNN stored using PB-BMP required $19.66\%$ more energy than CB for a VR distance of $3015$ using 5-bit weights. However, when trained with 2-bit weight precision, then for the same VR distance, PB-BMP can exploit the increased sparsity (27\%) in connectivity to consume $14.59\%$ lower energy than CB. %\cs{Also all claims in this paper are based on concrete energy numbers which previous works such as ~\cite{joshi2017neuromorphic, pedroni2019memory} have not provided.}

\bibliographystyle{IEEEtran}
\bibliography{biblio_unique_alt,iscas}

\end{document}